\documentclass{article}

\usepackage{algorithm}
\usepackage{algorithmic}
\usepackage{amsmath}
\usepackage{amsthm}
\usepackage[utf8]{inputenc}
\usepackage{microtype}
\usepackage{graphicx}
\usepackage{booktabs}
\usepackage{amsfonts}
\usepackage{amssymb}
\usepackage{nicefrac}
\usepackage{enumitem}
\usepackage{xcolor}

\definecolor{py-3-1}{HTML}{1F77B4}
\definecolor{py-3-2}{HTML}{FF7F0E}
\definecolor{py-3-3}{HTML}{2CA02C}
\definecolor{py-4-1}{HTML}{D62728}
\definecolor{py-4-2}{HTML}{9467BD}
\definecolor{py-4-3}{HTML}{BCBD22}
\definecolor{py-4-4}{HTML}{17BECF}

\newcommand{\F}{\mathbb{F}}
\renewcommand{\vec}[1]{{\bf #1}}
\newcommand{\mat}[1]{{\bf #1}}
\newcommand{\const}{\mathbf{c}}
\newcommand{\avec}{\mathbf{a}}
\newcommand{\x}{\mathbf{x}}
\newcommand{\y}{\mathbf{y}}
\newcommand{\z}{\mathbf{z}}
\newcommand{\A}{\mathbf{A}}
\newcommand{\X}{\mathbf{X}}
\newcommand{\Z}{\mathbf{Z}}
\newcommand{\btheta}{{\boldsymbol{\theta}}}
\newcommand{\blambda}{{\boldsymbol{\lambda}}}
\newcommand{\mul}{\texttt{mul}}
\newcommand{\share}[2]{\langle {#1}\rangle_{#2}}
\newcommand{\bZ}{\mathbb{Z}}
\newcommand{\bR}{\mathbb{R}}
\newcommand{\user}{{\textsc U}}
\newcommand{\dc}{\textsc{M}}
\newcommand{\reg}{\textsc{Reg}}

\newtheorem{proposition}{Proposition}

\PassOptionsToPackage{hyphens}{url}\usepackage{hyperref} 

\usepackage[accepted]{icml2018}

\icmltitlerunning{Blind Justice: Fairness with Encrypted Sensitive Attributes}

\begin{document}

\twocolumn[
\icmltitle{Blind Justice: Fairness with Encrypted Sensitive Attributes}



\icmlsetsymbol{equal}{*}

\begin{icmlauthorlist}
\icmlauthor{Niki Kilbertus}{max1,cam}
\icmlauthor{Adri\`{a} Gasc\'{o}n}{tur,war}
\icmlauthor{Matt Kusner}{tur,war}
\icmlauthor{Michael Veale}{ucl}
\icmlauthor{Krishna P. Gummadi}{max2}
\icmlauthor{Adrian Weller}{cam,tur}
\end{icmlauthorlist}

\icmlaffiliation{cam}{University of Cambridge}
\icmlaffiliation{tur}{The Alan Turing Institute}
\icmlaffiliation{war}{University of Warwick}
\icmlaffiliation{ucl}{University College London}
\icmlaffiliation{max1}{Max Planck Institute for Intelligent Systems}
\icmlaffiliation{max2}{Max Planck Institute for Software Systems}

 \icmlcorrespondingauthor{Niki Kilbertus}{niki.kilbertus@tuebingen.mpg.de}
\icmlkeywords{fairness,privacy,crypto,encrypted,fair learning,blind justice}

\vskip 0.3in
]



\printAffiliationsAndNotice{}  

\begin{abstract}
Recent work has explored how to train machine learning models which do not discriminate against any subgroup of the population as determined by sensitive attributes such as gender or race.
To avoid disparate treatment, sensitive attributes should not be considered.
On the other hand, in order to avoid disparate impact, sensitive attributes must be examined---e.g., in order to learn a fair model, or to check if a given model is fair.
We introduce methods from secure multi-party computation which allow us to avoid both.
By encrypting sensitive attributes, we show how an outcome-based fair model may be learned, checked, or have its outputs verified and held to account, \emph{without users revealing their sensitive attributes}.
\end{abstract}

\section{Introduction}
\label{sec:intro}

Concerns are rising that machine learning systems which make or support important decisions affecting individuals---such as car insurance pricing, r\'{e}sum\'{e} filtering or recidivism prediction---might illegally or unfairly discriminate against certain subgroups of the population \cite{Schreurs2008,Calders2012,barocas2016big}.
The growing field of \emph{fair learning} seeks to formalize relevant requirements, and through altering parts of the algorithmic decision-making pipeline, to detect and mitigate potential discrimination \cite{Friedler2016}.

Most legally-problematic discrimination centers on differences based on \emph{sensitive attributes}, such as gender or race \cite{barocas2016big}.
The first type, \emph{disparate treatment} (or \emph{direct discrimination}), occurs if individuals are treated differently according to their sensitive attributes (with all others equal).
To avoid disparate treatment, one should not inquire about individuals' sensitive attributes.
While this has some intuitive appeal and justification \cite{grgic2018}, a significant concern is that sensitive attributes may often be accurately predicted (``reconstructed'') from non-sensitive features \cite{Dwork}.
This motivates measures to deal with the second type of discrimination.

\emph{Disparate impact} (or \emph{indirect discrimination}) occurs when the \emph{outcomes} of decisions disproportionately benefit or hurt individuals from subgroups with particular sensitive attribute settings without appropriate justification.
For example, firms deploying car insurance telematics devices \cite{handel2014insurance} build up high dimensional pictures of driving behavior which might easily proxy for sensitive attributes even when they are omitted.
Much recent work in fair learning has focused on approaches to avoiding various notions of disparate impact \cite{FeldmanFMSV15,Hardtetal,zafar_fairness}.

In order to check and enforce such requirements, the modeler must have access to the sensitive attributes for individuals in the training data---however, this may be undesirable for several reasons~\cite{Zliobaite2016}.
First, individuals are unlikely to want to entrust sensitive attributes to modelers in all application domains.
Where applications have clear discriminatory potential, it is understandable that individuals may be wary of providing sensitive attributes to modelers who might exploit them to negative effect, especially with no guarantee that a fair model will indeed be learned and deployed.
Even if certain modelers themselves were trusted, the wide provision of sensitive data creates heightened privacy risks in the event of a data breach.

Further, legal barriers may limit collection and processing of sensitive personal data.
A timely example is the EU's General Data Protection Regulation (GDPR), which contains heightened prerequisites for the collection and processing of some sensitive attributes. 
Unlike other data, modelers cannot justify using sensitive characteristics in fair learning with their ``legitimate interests''---and instead will often need explicit, freely given consent~\cite{vealeedwardsa29}.

One way to address these concerns was recently proposed by \citet{VealeBinns2017}.
The idea is to involve a highly trusted third party, and may work well in some cases.
However, there are significant potential difficulties: individuals must disclose their sensitive attributes to the third party (even if an individual trusts the party, she may have concerns that the data may somehow be obtained or hacked by others, e.g.,~\citealp{Graham17});
and the modeler must disclose their model to the third party, which may be incompatible with their intellectual property or other business concerns.

\paragraph{Contribution.}
We propose an approach to detect and mitigate disparate impact without disclosing readable access to sensitive attributes.
This reflects the notion that decisions should be blind to an individual's status---depicted in courtrooms by a blindfolded Lady Justice holding balanced scales \cite{capers2012blind}.
We assume the existence of a regulator with fairness aims (such as a data protection authority or anti-discrimination agency).
With recent methods from \emph{secure multi-party computation} (MPC), we enable auditable fair learning while ensuring that both individuals' sensitive attributes and the modeler's model remain private to all other parties---including the regulator.
Desirable fairness and accountability applications we enable include:

\begin{enumerate}[leftmargin=1em,topsep=0pt,itemsep=0ex,partopsep=1ex,parsep=1ex]
\item \textbf{Fairness certification.} Given a model and a dataset of individuals, check that the model satisfies a given fairness constraint (we consider several notions from the literature, see Section~\ref{sec:fairnesscrit}); if yes, generate a certificate.
\item \textbf{Fair model training.} Given a dataset of individuals, learn a model guaranteed and certified to be fair.
\item \textbf{Decision verification.} A malicious modeler might go through fair model training, but then use a different model in practice.
To address such accountability concerns \cite{kroll2016accountable}, we efficiently provide for an individual to challenge a received outcome, verifying that it matches the outcome from the previously certified model.
\end{enumerate}

We rely on recent theoretical developments in MPC (see Section~\ref{sec:mpc}) which we extend to admit linear constraints in order to enforce fairness requirements.
These extensions may be of independent interest.
We demonstrate the real-world efficacy of our methods, and shall make our code publicly available.

\section{Fairness and Privacy Requirements}
\label{sec:setting}

Here we formalize our setup and requirements.

\subsection{Assumptions and Incentives}

We assume three categories of participants: a \emph{modeler}~$\dc$, a {\em regulator}~$\reg$, and \emph{users} $\user_1, \ldots, \user_n$.
For each user, we consider a vector of sensitive features (or attributes, we use the terms interchangeably) $\z_i\in\cal{Z}$ (e.g., ethnicity or gender) which might be a source of discrimination, and a vector of non-sensitive features $\x_i\in\cal{X}$ (discrete or real).
Additionally, each user has a non-sensitive feature $y_i\in\cal{Y}$ which the modeler~$\dc$ would like to predict---the \emph{label} (e.g., loan default). In line with current work in fair learning, we assume that all $\z_i$ and $y_i$ attributes are binary, though our MPC approach could be extended to multi-label settings.
The source of societal concern is that sensitive attributes~$\z_i$ are potentially correlated with~$\x_i$ or $y_i$.

Modeler~$\dc$ wishes to train a model $f_{\btheta}: {\cal X}\to {\cal Y}$, which accurately maps features~$\x_i$ to labels~$y_i$, in a supervised fashion.
We assume $\dc$ needs to keep the model private for intellectual property or other business reasons.
The model $f_\btheta$ does not use sensitive information $\z_i$ as input to prevent disparate treatment (direct discrimination).

For each user $\user_i$, $\dc$ observes or is provided $\x_i, y_i$.
The sensitive information in~$\z_i$ is required to ensure~$f_{\btheta}$ meets a given disparate impact fairness condition~$\F$ (see Section~\ref{sec:fairnesscrit}).
While each user~$\user_i$ wants~$f_{\btheta}$ to meet~$\F$, they also wish to keep~$\z_i$ private from all other parties. The regulator~$\reg$ aims to ensure that~$\dc$ deploys only models that meet fairness condition~$\F$. It has no incentive to collude with~$\dc$ (if collusion were a concern, more sophisticated cryptographic protocols would be required).
Further, the modeler~$\dc$ might be legally obliged to demonstrate to the regulator~$\reg$ that their model meets fairness condition~$\F$ before it can be publicly deployed.
As part of this,~$\reg$ also has a positive duty to enable the training of fair models.

In Section~\ref{sec:3problems}, we define and address three fundamental problems in our setup: certification, training, and verification.
For each problem, we present its functional goal and its privacy requirements.
We refer to $\mat{D} = \{ (\x_i, y_i) \}_{i=1}^{n}$ and $\Z = \{\z_i\}_{i=1}^{n}$ as the non-sensitive and sensitive data, respectively.
In Section~\ref{sec:fairnesscrit}, we first provide necessary background on various notions of fairness that have been explored in the fair learning literature.

\subsection{Fairness Criteria} \label{sec:fairnesscrit}
In large part, works that formalize fairness in machine learning do so by balancing a certain condition between groups of people with different sensitive attributes, $\z$ versus $\z'$.
Several possible conditions have been proposed.
Popular choices include (where $y \in \{0,1\}$ and $\hat{y}$ is the prediction of a machine learning model):
\begin{align}
P(\hat{y} = y \mid \z) &= P(\hat{y} = y \mid \z') & \!\!\textrm{(acc)} \label{eq:acc} \\
P(\hat{y} = y \mid \z, y = 1) &= P(\hat{y} = y \mid \z', y = 1) & \!\!\textrm{(TPR)} \label{eq:tpr} \\
P(\hat{y} = y \mid \z, y = 0) &= P(\hat{y} = y \mid \z', y = 0) & \!\!\textrm{(TNR)} \label{eq:tnr} \\
P(\hat{y} = y \mid \z, \hat{y}=1) &= P(\hat{y} = y \mid \z', \hat{y}=1) & \!\!\textrm{(PPV)} \label{eq:ppv} \\
P(\hat{y} = y \mid \z, \hat{y}=0) &= P(\hat{y} = y \mid \z', \hat{y}=0) & \!\!\textrm{(NPV)} \label{eq:npv} \\
P(\hat{y} = 1 \mid \z) &= P(\hat{y} = 1 \mid \z') & \!\!\textrm{(AR)} \label{eq:ar}
\end{align}
Respectively, these consider equality of: \eqref{eq:acc} accuracy, \eqref{eq:tpr} true positive rate, \eqref{eq:tnr} true negative rate, \eqref{eq:ppv} positive predicted value, \eqref{eq:npv} negative predicted value, or \eqref{eq:ar} acceptance rate.
Works which use these or related notions include \citep{Hardtetal,zafar_fairness,Zafaretal,zafar2017parity}.

In this work we focus on a variant of eq.~\eqref{eq:ar}, formulated as a constrained optimization problem by \citet{zafar_fairness} mimicking the~$p\%$-rule \citep{biddle2006adverse}: for any binary protected attribute~$z \in \{0,1\}$, it aims to achieve
\begin{equation}\label{eq:p_percent}
\min\!\left\{
\frac{P(\hat{y} = 1 \,|\, z=1)}{P(\hat{y} = 1 \,|\, z=0)},
\frac{P(\hat{y} = 1 \,|\, z=0)}{P(\hat{y} = 1 \,|\, z=1)}
\right\} \!\ge\! \frac{p}{100}\:.
\end{equation}

We believe that in future work, a similar MPC approach could also be used for conditions \eqref{eq:acc}, \eqref{eq:tpr} or \eqref{eq:tnr}---i.e., all the other measures which, to our knowledge, have been addressed with efficient standard (non-private) methods.

\begin{figure*}[t!]
  \centerline{\includegraphics[width=0.85\textwidth]{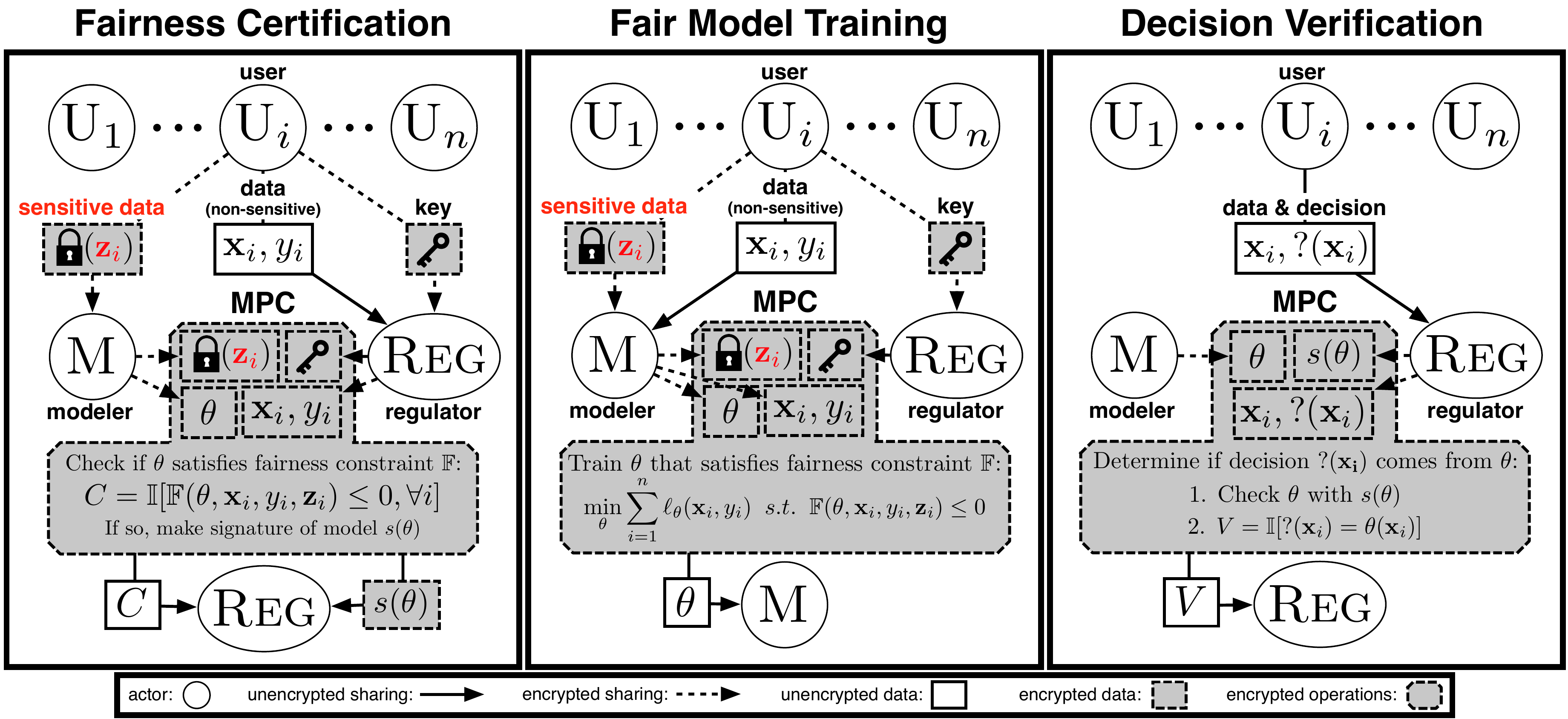}}
  \vspace{-2ex}
  \caption{Our setup for \emph{Fairness certification} (\emph{Left}), \emph{Fair model training} (\emph{Center}), and \emph{Decision verification} (\emph{Right}).}
  \label{figure.MPC}
  \vspace{-0.3cm}
\end{figure*}

\subsection{Certification, Training, and Verification}\label{sec:3problems}

\paragraph{Fairness certification.}
Given a notion of fairness $\F$, the modeler $\dc$ would like to work with the regulator $\reg$ to obtain a certificate that model $f_{\btheta}$ is fair.
To do so, we propose that users send their non-sensitive data $\mat{D}$ to $\reg$;
and send \emph{encrypted} versions of their sensitive data $\Z$ to both $\dc$ and $\reg$.
Neither $\dc$ nor $\reg$ can read the sensitive data.
However, we can design a secure protocol between $\dc$ and $\reg$ (described in Section~\ref{sec:mpc}) to certify if the model is fair.
This setup is shown in Figure~\ref{figure.MPC} (\emph{Left}).

While both~$\reg$ and~$\dc$ learn the outcome of the certification, we require the following \emph{privacy constraints}:
(C1) \emph{privacy of sensitive user data:} no one other than $\user_i$ ever learns $\z_i$ in the clear,
(C2) \emph{model secrecy:} only $\dc$ learns $f_{\btheta}$ in the clear, and
(C3) \emph{minimal disclosure of $\mat{D}$ to $\reg$:} only $\reg$ learns $\mat{D}$ in the clear.

\paragraph{Fair model training.}
How can a modeler $\dc$ learn a fair model without access to users' sensitive data $\Z$? We propose to solve this by having users send their non-sensitive data $\mat{D}$ to $\dc$ and to distribute encryptions of their sensitive data to $\dc$ and $\reg$ as in certification.
We shall describe a secure MPC protocol between $\dc$ and $\reg$ to train a fair model $f_{\btheta}$ privately. This setup is shown in Figure~\ref{figure.MPC} (\emph{Center}).

\emph{Privacy constraints}:
(C1) privacy of sensitive user data,
(C2) model secrecy, and
(C3) minimal disclosure of $\mat{D}$ to $\dc$.

\paragraph{Decision verification.}
Assume that a malicious $\dc$ has had model~$f_{\btheta}$ successfully certified by $\reg$ as above.
It then swaps $f_{\btheta}$ for another potentially unfair model $f_{\btheta'}$ in the real world.
When a user receives a decision $\hat{y}$, e.g., her mortgage is denied, she can then challenge that decision by asking~$\reg$ for a verification $V$.
The verification involves $\dc$ and $\reg$, and consists of verifying that $f_{\btheta'}(\x) = f_{\btheta}(\x)$, where $\x$ is the user's non-sensitive data.
This ensures that the user would have been subject to the same result with the certified model $f_{\btheta}$, even if $f_{\btheta'} \neq f_{\btheta}$ and $f_{\btheta'}$ is not fair.
Hence, while there is no simple technical way to prevent a malicious $\dc$ from deploying an unfair model, it will get caught if a user challenges a decision that would differ under~$f_{\btheta}$. This setup is shown in Figure~\ref{figure.MPC} (\emph{Right}).

\emph{Privacy constraint}: While $\reg$ and the user learn the outcome of the verification, we require
(C1) privacy of sensitive user data, and
(C2) model secrecy.

\subsection{Design Choices} \label{sec:designchoices}

We use a regulator for several reasons. Given fair learning is of most benefit to vulnerable individuals, we do not wish to deter adoption with high individual burdens.
While MPC could be carried out without the involvement of a regulator, using all users as parties, this comes at a significantly greater computational cost.
With current methods, taking that approach would be unrealistic given the size of the user-base in many domains of concern, and would furthermore require all users to be online simultaneously.
Introducing a regulator removes these barriers and leaves users' computational burden at a minimum level, with envisaged applications practical with only their web browsers.

In cases where users are uncomfortable sharing $\mat{D}$ with either $\reg$ or $\dc$, it is trivial to extend all three tasks such that all of $\x_i, y_i,\z_i$ remain private throughout, with the computation cost increasing only by a factor of 2.
This extension would sometimes be desirable as it restricts the view of $\dc$ to the final model, prohibiting inferences about $\Z$ when $\mat{D}$ is known.
However, this setup hinders exploratory data analysis by the modeler which might promote robust model-building, and, in the case of verification, validation by the regulator that user-provided data is correct.

\section{Our Solution}
\label{sec:mpc}

Our proposed solution to these three problems is to use Multi-Party Computation (MPC).
Before we describe how it can be applied to fair learning, we first present the basic principles of MPC, as well as its limitations particularly in the context of machine learning applications.

\subsection{MPC for Machine Learning} \label{sec:MPC}

Multi-Party Computation protocols allow two parties~$P_1$ and~$P_2$ holding secret values~$x_1$ and~$x_2$ to evaluate an agreed-upon function $f$, via $y = f(x_1, x_2)$ in a way in which the parties (either both or one of them) learn \emph{only}~$y$.
For example, if $f(x_1, x_2) = \mathbb{I}(x_1 < x_2)$, then the parties would learn which of their values is bigger, but nothing else.\footnote{The function $\mathbb{I}$ is $1$ if its argument is true and $0$ otherwise.}
This corresponds to the well-known \emph{Yao's millionaires problem}: two millionaires want to conclude who is richer without disclosing their wealth to each other.
The problem was introduced by Andrew Yao in 1982, and kicked off the area of multi-party computation in cryptography.

In our setting---instead of a simple comparison as in the millionaires problem---$f$ will be either
(i) a procedure to check the fairness of a model and certify it,
(ii) a machine learning training procedure with fairness constraints, or
(iii) a model evaluation to verify a decision.
The two parties involved in our computation are the modeler $\dc$ and the regulator $\reg$.
The inputs depend on the case (see Figure~\ref{figure.MPC}).

As generic solutions do not yet scale to real-world data analysis tasks, one typically has to tailor custom protocols to the desired functionality.
This approach has been followed successfully for a variety of machine learning tasks such as logistic and linear regression \cite{nikolaenko_privacy-preserving_2013, gascon_privacy-preserving_2017, mohassel2017secureml}, neural network training \cite{mohassel2017secureml} and evaluation \cite{gazelle, minionn}, matrix factorization \cite{nikolaenko_matrix-factorisation}, and principal component analysis \cite{pca}.
In the next section we review challenges beyond scalability issues that arise when implementing machine learning algorithms in MPC.

\subsection{Challenges in Multi-Party Machine Learning}

MPC protocols can be classified into two groups depending on whether the target function is represented as either a Boolean or arithmetic circuit.
All protocols proceed by having the parties jointly evaluate the circuit, processing it gate by gate while keeping intermediate values hidden from both parties by means of a secret sharing scheme.
While representing functions as circuits can be done without losing expressiveness, it means certain operations are impractical.
In particular, algorithms that execute different branches depending on the input data will explode in size when implemented as circuits, and in some cases lose their run time guarantees (e.g., consider binary search).

Crucially, this applies to \emph{floating-point arithmetic}.
While this is work in progress, state-of-the-art MPC floating-point arithmetic implementations take more than~$15$ milliseconds to multiply two $64$ bit numbers \citep[Table~4]{DDKSSZ15}, which is prohibitive for our applications.
Hence, machine learning MPC protocols are limited to \emph{fixed-point} arithmetic.
Overcoming this limitation is a key challenge for the field.
Another necessity for the feasibility of MPC is to approximate non-linear functions such as the sigmoid, ideally by (piecewise) linear functions.

\subsection{Our MPC Protocols}

\paragraph{Input sharing.}
To implement the functionality from Figure~\ref{figure.MPC}, we first need a  secure procedure for the users to \emph{secret share} a sensitive value, for example her race, with the modeler~$\dc$ and the regulator~$\reg$.
We use \emph{additive secret sharing}.
A value~$z$ is represented in a finite domain~$\bZ_q$---we use~$q=2^{64}$.
To share~$z$, the user samples a value~$r$ from~$\bZ_q$ uniformly at random, and sends $z - r$ to~$\dc$ and~$r$ to~$\reg$.
While $z$ can be reconstructed (and subsequently operated on) inside the MPC computation by means of a simple addition, each share on its own does not reveal anything~$z$ (other than that it is in~$\bZ_q$).
One can think of arithmetic sharing as a ``distributed one-time pad''.

In Figure~\ref{figure.MPC}, we now reinterpret the key held by $\reg$ and the encrypted $z$ by $\dc$, as their corresponding shares of the sensitive attributes and denote them by $\share{z}{1}$ and $\share{z}{2}$ respectively.
The idea of privately outsourcing computation to two non-colluding parties in this way is recurrent in MPC, and often referred to as the two-server model~\cite{mohassel2017secureml, gascon_privacy-preserving_2017, nikolaenko_privacy-preserving_2013, pca}.

\paragraph{Signing and checking a model.}
Note that \emph{certification} and \emph{verification} partly correspond to sub-procedures of the \emph{fair training} task:
during training we check the fairness constraint $\F$, and repeatedly evaluate partial models on the training dataset (using gradient descent).
Hence, \emph{certification} and \emph{verification} do not add technical difficulties over training, which is described in detail in Section~\ref{sec:tc}.
However, for verification, we still need to ``sign'' the model, i.e., $\reg$ should obtain a signature $s(\btheta)$ as a result of model certification, see Figure~\ref{figure.MPC} (\emph{Left}).
This signature is used to check in the verification phase, whether a given model $\btheta'$ from $\dc$ satisfies $s(\btheta') = s(\btheta)$ for a certified fair model $\btheta$ (in which case $\btheta = \btheta'$
with high probability).
Moreover, we need to preserve the secrecy of the model, i.e., $\reg$ should not be able to recover $\btheta$ from $s(\btheta)$.
These properties, given that the space of models is large, calls for a cryptographic hash function, such as SHA-256.

Additionally, in our functionality, the hash of $\btheta$ should be computed inside MPC, to hide $\btheta$ from $\reg$.
Fortunately, cryptographic hashes such as SHA-256 are a common benchmark functionality in MPC, and their execution is highly optimized.
More concretely, the overhead of computing $s(\btheta)$, which needs to be done both for certification and verification is of the order of fractions of a second \citep[Figure~14]{sha-mpc}.
While cryptographic hash functions have various applications in MPC, we believe the application to machine learning model certification is novel.

Hence, certification is implemented in MPC as a check that $\btheta$ satisfies the criterion $\F$, followed by the computation of $s(\btheta)$.
On the other hand, for verification, the MPC protocol first computes the signature of the model provided by $\dc$, and then proceeds with a prediction as long as the computed signature matches the one obtained by $\reg$ in the verification phase.
An alternative solution is possible based on symmetric encryption under a shared key, as highly efficient MPC implementations of block ciphers such as AES are available~\cite{KellerORSSV17}.

\paragraph{Fair training.}
To realize the \emph{fair training} functionality from the previous section, we follow closely the techniques recently introduced by \citet{mohassel2017secureml}.
Specifically, we extend their custom MPC protocol for logistic regression to additionally handle linear constraints.
This extension may be of independent interest, and has applications for privacy-preserving machine learning beyond fairness.
The concrete technical difficulties in achieving this goal, and how to overcome them, are presented in the next section.
The formal privacy guarantees of our fair training protocol are stated in the following proposition.

\begin{proposition}
For non-colluding~$\dc$ and~$\reg$, our protocol implements the fair model training functionality satisfying constraints (C1)-(C3) in Section~\ref{sec:3problems} in the presence of a semi-honest adversary.
\end{proposition}
The proof holds in the random oracle model, as a standard simulation argument combining several MPC primitives \cite{mohassel2017secureml,gascon_privacy-preserving_2017}.
It leverages security of arithmetic sharing, garbled circuits, and oblivious transfer protocols in the semi-honest model \cite{DBLP:conf/stoc/GoldreichMW87}.
A general introduction to MPC, as well as a description of the relevant techniques from~\cite{mohassel2017secureml} used in our protocol, can be found in Section~\ref{sec:mpcdetails} in the appendix.

\section{Technical Challenges of Fair Training}
\label{sec:tc}

We now present our tailored approaches for learning and evaluating fair models with encrypted sensitive attributes.
We do this via the following contributions:
\begin{itemize}[leftmargin=*, noitemsep, topsep=0pt]
\item We argue that current optimization techniques for fair learning algorithms are unstable for fixed-point data, which is required by our MPC techniques.
\item We describe optimization schemes that are well-suited for learning over fixed-point number representations.
\item We combine tricks to approximate non-linear functions with specialized operations to make fixed-point arithmetic feasible and avoid over- and under-flows.
\end{itemize}

The optimization problem at hand is to learn a classifier $\btheta$ subject to a (often convex) fairness constraint~$\F(\btheta)$:
\begin{equation}\label{eq:optim}
\min_\btheta\;\; \sum_{i=1}^{n}\ell_{\btheta}(\x_i, y_i)\qquad
\text{subject to}\;\; \F(\btheta) \le \mathbf{0}\:,
\end{equation}
where $\ell_{\btheta}$ is a loss term (the logistic loss in this work).
We collect user data from~$\user_1, \ldots, \user_n$ into matrices~$\X \in \bR^{n\times d}, \Z\in\{0,1\}^{n\times p}$ and a label vector~$\y \in \{0,1\}^n$.

\citet{zafar_fairness} use a convex approximation of the $p\%$-rule, see eq.~\eqref{eq:p_percent}, for linear classifiers to derive the constraint:
\begin{equation}\label{eq:ppercent}
\F(\btheta) = \frac{1}{n} |\hat{\Z}^{\top} \X \btheta| - \const \:,
\end{equation}
where~$\hat{\Z}$ is the matrix of all $\hat{\z}_i : = \z_i - \bar{\z}$ and~$\const \in \bR^d$ is a constant vector corresponding to the tightness of the fairness constraint. Here, $\bar{\z}$ is the mean of all inputs $\z_i$. With $\A := \nicefrac{1}{n} \hat{\Z}^{\top} \X$, the \emph{$p\%$ constraint} reads~$\F(\btheta) = |\A \btheta | - \const$, where the absolute value is taken element-wise.

\subsection{Current Techniques}\label{subsec:current}

To solve the optimization problem in eq.~\eqref{eq:optim}, with the fairness function~$\F{}$ in eq.~\eqref{eq:ppercent}, \citet{zafar_fairness} use Sequential Least Squares Programming (SLSQP).
This technique works by reformulating eq.~\eqref{eq:optim} as a sequence of Quadratic Programs (QPs).
After solving each QP, their algorithm uses the Han-Powell method, a quasi-Newton method that iteratively approximates the Hessian $\mathbf{H}$ of the objective function via the update
\begin{align}
\mathbf{H}_{t+1} = \mathbf{H}_t + \frac{\mathbf{l}_{\Delta} \mathbf{l}_{\Delta}^\top}{\mathbf{\btheta}_{\Delta}^\top \mathbf{l}_{\Delta}} - \frac{\mathbf{H}_t \mathbf{\btheta}_{\Delta} \mathbf{\btheta}_{\Delta}^\top \mathbf{H}_t}{\mathbf{\btheta}_{\Delta}^\top \mathbf{H}_t\mathbf{\btheta}_{\Delta}} \:, \nonumber
\end{align}
where $\mathbf{l}_{\Delta} = \mathbf{l}(\btheta_{t+1},\boldsymbol{\lambda}_{t+1}) - \mathbf{l}(\btheta_{t},\boldsymbol{\lambda}_{t})$ and $\mathbf{l}(\btheta_t,\boldsymbol{\lambda}_t) = \sum_{i=1}^n \ell_{\btheta_t}(\mathbf{x}_i,y_i) + \boldsymbol{\lambda}^\top \F(\btheta_t)$ is the Lagrangian of eq.~\eqref{eq:optim}.
Finally, $\mathbf{\btheta}_{\Delta} = \mathbf{\btheta}_{t+1} - \mathbf{\btheta}_t$.

There are two issues with this approach from an MPC perspective.
First, solving a sequence of QPs is prohibitively time-consuming in MPC.
Second, while the above Han-Powell update performs well on floating-point data, the two divisions by non-constant, non-integer numbers easily underflow or overflow with fixed-point numbers.

\subsection{Fixed-Point-Friendly Optimization Techniques}

Instead, to solve the optimization problem in eq.~\eqref{eq:optim}, we perform stochastic gradient descent and experiment with the following techniques to incorporate the constraints.

\emph{Lagrangian multipliers.}
Here we minimize
\begin{equation*}
\mathcal{L} := \frac{1}{n} \sum_{i=1}^{n} \ell_{\btheta}^{\mathrm{BCE}}(\x_i, y_i) + \blambda^{\top} \max\{ \F(\btheta), \mathbf{0} \} \:,
\end{equation*}
using stochastic gradient descent, i.e., alternating updates
$\btheta \gets \btheta - \eta_{\btheta} \nabla_{\btheta} \mathcal{L}$ and $\blambda \gets \max\{\blambda + \eta_{\blambda} \nabla_{\blambda} \mathcal{L}, \mathbf{0} \}$,
where~$\eta_{\btheta}, \eta_{\blambda}$ are the learning rates.

\emph{Projected gradient descent.}
For this method, consider specifically the~$p\%$-rule based notion~$\F(\btheta) = |\A \btheta | - \const$. We first define~$\hat{\A}$ as the matrix consisting of the rows of~$\A$ for which~$\F(\btheta) > \mathbf{0}$, i.e., where the constraint is active.
In each step, we project the computed gradient of the binary-cross-entropy loss~$\mathcal{L}^{\mathrm{BCE}}$---either of a single example or averaged over a minibatch---back into the constraint set, i.e.,
\begin{equation}\label{eq:projected}
\btheta \gets \btheta - \eta_{\btheta} (\mathrm{Id}_d - \hat{\A}^{\top} (\hat{\A} \hat{\A}^{\top})^{-1} \hat{\A}) \nabla_{\btheta} \ell_{\btheta}^{\mathrm{BCE}}\:.
\end{equation}

\emph{Interior point log barrier~\cite{boydsbook}.}
We can approximate eq.~\eqref{eq:optim} for the~$p\%$-rule constraint~$\F(\btheta) = |\A \btheta | - \const$ by:
minimize $\sum_{i=1}^{n} \ell_{\btheta}^{\mathrm{BCE}}(\x_i, y_i)
- \frac{1}{t} \sum_{j=1}^{p}
\bigl(\log(\avec_j^{\top} \btheta + c_j) + \log(-\avec_j^{\top} \btheta + c_j)\bigr)$, where~$\avec_j$ is the~$j$th row of~$\A$.
The parameter~$t$ trades off the approximation of the true objective ($I_{-}(u) = 0$ for~$u \le 0$ and $I_{-}(u) = \infty$ for~$u > 0$) and the smoothness of the objective function.
Throughout training~$t$ is increased, allowing the solution to move closer to the boundary.
As the gradient of the objective has a simple closed form representation, we can perform regular (stochastic) gradient descent.

After extensive experiments (see Section~\ref{sec:experiments}) we found the Lagrangian multipliers technique to work best, both in yielding high accuracies, reliably staying within the constraints and being robust to hyperparameter changes such as learning rates or the batch size.
For a proof of concept, in Section~\ref{sec:experiments} we focus on the $p\%$-rule, i.e.,~eq.~\eqref{eq:ppercent}.
Note that the gradients for eq.~\eqref{eq:tpr} and eq.~\eqref{eq:tnr} take a similarly simple form, i.e., balancing the true positive or true negative rates (corresponding to equal opportunity or equal odds) is simple to implement for the Lagrangian multiplier technique, but harder for projected gradient descent.
However, these fairness notions are more expensive as we have to compute~$\Z^{\top} \X$ for each update step, instead of pre-computing it once at the beginning of training, see Algorithm~\ref{algo:lagrange} in the appendix.
We could speed up the computation again by evaluating the constraint only on the current minibatch for each update, in which case we risk violating the fairness constraint.

\paragraph{MPC-friendliness.}
For eq.~\eqref{eq:ppercent}, we can compute the gradient updates in all three methods with elementary linear algebra operations (matrix multiplications) and a single evaluation of the logistic function.
While MPC is well suited for linear operations, most nonlinear functions are prohibitively expensive to evaluate in an MPC framework.
Hence we tried two piecewise linear approximations for~$\sigma(x)$.
The first was recently suggested for machine learning in an MPC context~\citep{mohassel2017secureml} and is simply constant~$0$ and~$1$ for $x< -0.5$ and $x>0.5$ respectively, and linear in between.
The second uses the optimal first order Chebychev polynomial on each interval~$[x, x+1]$ for~$x \in \{-5, -4, \ldots, 4\}$, and is constant~$0$ or~$1$ outside of~$[-5,5]$ \citep{sigmoid-approx}.
While it is more accurate, we only report results for the simpler first approximation, as it yielded equal or better results in all our experiments.

As the largest number that can be represented in fixed-point format with~$m$ integer and~$m$ fractional bits is roughly~$2^m + 1$, overflow becomes a common problem.
Since we whiten the features~$\X$ column-wise, we need to be careful whenever we add roughly~$2^m$ numbers or more, because we cannot even represent numbers greater than $2^m$.
In particular, the minibatch size has to be smaller than this limit.
For large~$n$, the multiplication~$\Z^{\top} \X$ in the fairness function~$\F$ for the $p\%$-rule is particularly problematic.

Hence, we split both factors into blocks of size $b\times b$ with~$b < 2^m$ and normalize the result of each blocked matrix multiplication by~$b$ before adding up the blocks.
We then multiply the sum by~$\nicefrac{b}{n} > 2^{-m}$.
As long as~$b$, $\nicefrac{b}{n}$ (and thus also~$\nicefrac{n}{b}$) can be represented with sufficient precision, which is the case in all our experiments, this procedure avoids under- and overflow.
Note that we require the sample size~$n$ to be a multiple of~$b$.
In practice, we have to either discard or duplicate part of the data.
Since the latter may introduce bias, we recommend subsampling.
Once we have (an approximation of) $\A \in \bR^{p \times d}$, we resort to normal matrix multiplication, as typically~$p, d \lesssim 100$, see Table~\ref{table.timing}.

Division is prohibitively expensive in MPC.
Hence, we set the minibatch size to a power of two, which allows us to use fast bit shifts for divisions when averaging over minibatches.
To exploit the same trick when averaging over/across blocks in the blocked matrix multiplication, we choose~$n$ as the largest possible power of two, see Table~\ref{table.timing}.
Algorithm~\ref{algo:lagrange} in Section~\ref{sec:algo} in the appendix describes the computations $\dc$ and $\reg$ have to run for fair model training using the Lagrangian multiplier technique and the $p\%$-rule from eq.~\eqref{eq:ppercent}.
We implicitly assume all computations are performed jointly on additively shared secrets.

\section{Experiments}
\label{sec:experiments}

\begin{table}
\centering
\caption{Dataset sizes and online timing results of MPC certification and training over 10 epochs with batch size~64.}
\label{table.timing}

\begin{scriptsize}
\begin{tabular}{lrrrrr}
\toprule
 &  Adult & Bank & COMPAS & German & SQF  \\
  \midrule
$n$ training examples & $2^{14}$ & $2^{15}$ & $2^{12}$ & $2^9$ & $2^{16}$
\\
$d$ features & 51 & 62 & 7 & 24 & 23
\\
$p$ sensitive attr.& 1 & 1 & 7 & 1 & 1
\\
certification & 802~ms & 827~ms & 288~ms & 250~ms & 765~ms
\\
training & 43~min & 51~min & 7~min & 1~min & 111~min \\
\bottomrule
\end{tabular}
\end{scriptsize}
\vspace{-0.6cm}
\end{table}

The root cause for most technical difficulties pointed out in the previous section is the necessity to work with fixed-point numbers and the high computational cost of MPC.
Hence, major concerns are loss of precision and infeasible running times.
In this section, we show how to overcome both doubts and that fair training, certification and verification are feasible for realistic datasets.

\subsection{Experimental Setup and Datasets}

We work with two separate code bases.
Our Python code does not implement MPC, to be able to flexibly switch between floating and fixed-point numbers as well as exact non-linear functions and their approximations.
We use it mostly for validation and empirical guidance in our design choices.
The full MPC protocol is implemented in C++ on top of the Obliv-C garbled circuits framework~\cite{zahur2015obliv} and the Absentminded Crypto Kit~\cite{liback}. This is done as described in Section~\ref{sec:mpc} for the Lagrangian multiplier technique (see Section~\ref{sec:mpcdetails} in the appendix for more details).
It accurately mirrors the computations performed by the first implementation on encrypted data.\footnote{Code is available at \url{https://github.com/nikikilbertus/blind-justice}}
Except for the timing results in Table~\ref{table.timing}, all comparisons with floating-point numbers or non-linearities were done with the versatile Python implementation.
Details about parameters and the algorithm can be found in Section~\ref{sec:algo} in the appendix.

\begin{figure*}
\centering
\includegraphics{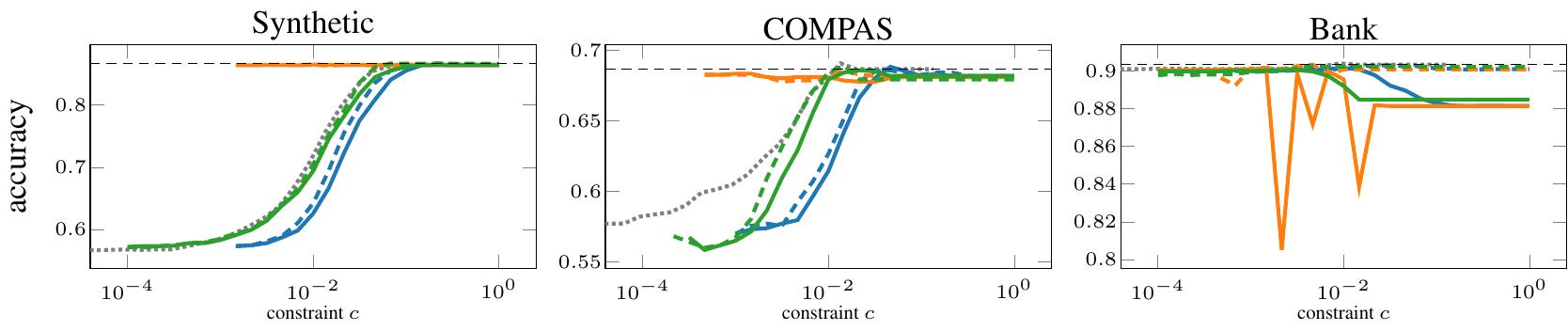}
\vspace{-0.4cm}
\caption{Test set accuracy over the $p\%$ value for different optimization methods ({\color{py-3-1} blue: iplb}, {\color{py-3-2} orange: projected}, {\color{py-3-3} green: Lagrange}) and either no approximation (\emph{continuous}) or a piecewise linear approximation (\emph{dashed}) of the sigmoid using floating-point numbers. The gray dotted line is the baseline (see Section~\ref{subsec:current}) and the black dashed line is unconstrained logistic regression (from scikit-learn).}
\label{fig:ppercent_acc}
\vspace{-0.3cm}
\end{figure*}

\begin{figure*}
\centering
\includegraphics{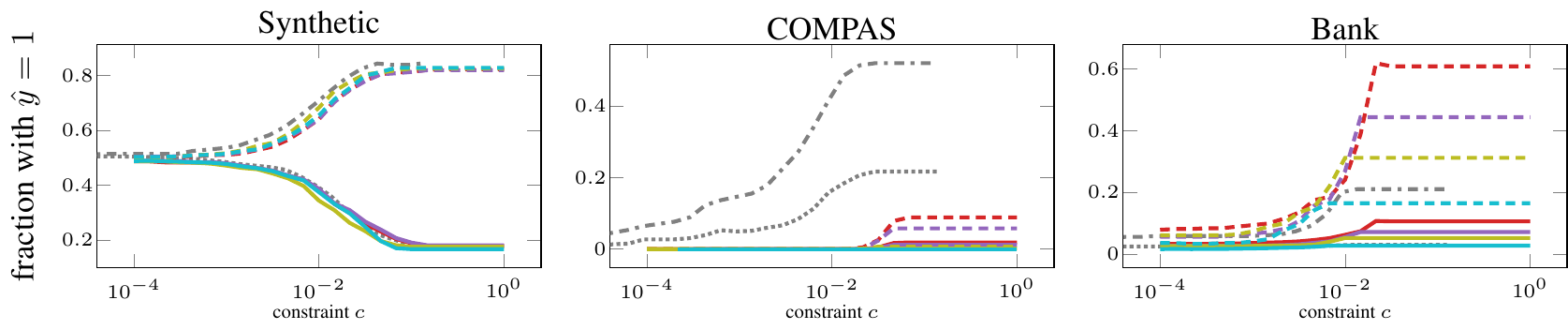}
\vspace{-0.4cm}
\caption{The fraction of people with $z=0$ (\emph{continuous/dotted}) and $z=1$ (\emph{dashed/dash-dotted}) who get assigned positive outcomes
({\color{py-4-1}red: no approx. + float},
{\color{py-4-2}purple: no approx. + fixed},
{\color{py-4-3}yellow: pw linear + float},
{\color{py-4-4}turquoise: pw linear + fixed},
{gray: baseline}).}
\label{fig:ppercent_fair}
\vspace{-0.3cm}
\end{figure*}

We consider~5 real world datasets, namely the adult (\emph{Adult}), German credit (\emph{German}), and bank market (\emph{Bank}) datasets from the UCI machine learning repository \citep{uci}, the stop, question and frisk 2012 dataset (\emph{SQF}),\footnote{\url{https://perma.cc/6CSM-N7AQ}} and the COMPAS dataset \cite{Angwin2016} (\emph{COMPAS}).
For practical purposes (see Section~\ref{sec:tc}), we subsample~$2^i$ examples from each dataset with the largest possible~$i$, see
Table~\ref{table.timing}.
Moreover, we also run on synthetic data, generated as described by~\citet[Section 4.1]{zafar_fairness}, as it allows us to control the correlation between the sensitive attributes and the class labels.
It is thus well suited to observe how different optimization techniques handle the fairness-accuracy trade off.
For comparison we use the SLSQP approach described in Section~\ref{subsec:current} as a baseline.
We run all methods for a range of constraint values in~$[10^{-4}, 10^0]$ and a corresponding range for SLSQP.

In the plots in this section, discontinuations of lines indicate failed experiments.
The most common reasons are overflow and underflow for fixed-point numbers, and instability due to exploding gradients.
Plots and analyses for the remaining datasets can be found in Section~\ref{sec:additional} in the appendix.

\subsection{Comparing Optimization Techniques}

First we evaluate which of the three optimization techniques works best in practice.
Figure~\ref{fig:ppercent_acc} shows the test set accuracy over the constraint value.
By design, the synthetic dataset exhibits a clear trade-off between accuracy and fairness.
The {\color{py-3-3}Lagrange} technique closely follows the (dotted) baseline from \cite{zafar_fairness}, whereas {\color{py-3-1}iplb} performs slightly worse (and fails for small $c$).
Even though the {\color{py-3-2}projected} gradient method formally satisfies the proxy constraint for the $p\%$ rule, it does so by merely shrinking the parameter vector $\btheta$, which is why it also fails for small $c$.
We analyze this behavior in more detail in Section~\ref{sec:additional} in the appendix.

The COMPAS dataset is the most challenging as it contains 7 sensitive attributes, one of which has only 10 positive instances in the training set.
Since we enforce the fairness constraint individually for each sensitive attribute (we randomly picked one for visualization), the  classifier tends to collapse to negative predictions.
All three methods maintain close to optimal accuracy in the unconstrained region, but collapse more quickly than SLSQP.
This example shows that the $p\%$-rule proxy itself needs careful interpretation when applied to multiple sensitive attributes simultaneously and that our SGD based approach seems particularly prone to collapse in such a scenario.
On the Bank dataset accuracy increases for {\color{py-3-1}iplb} and {\color{py-3-3}Lagrange} when the constraint becomes active as $c$ decreases until they match the baseline.
Determining the cause of this---perhaps unintuitive---behavior requires further investigation.
We currently suspect the constraint to act as a regularizer.
The {\color{py-3-2}projected} gradient method is unreliable on the Bank dataset.

Empirically, the Lagrangian multiplier technique is most robust with maximal deviations of accuracy from SLSQP of $<4$\% across the 6 datasets and all constraint values.
We substantiate this claim in Section~\ref{sec:additional} of the appendix.
For the rest of this section we only report results for Lagrangian multipliers.
Figure~\ref{fig:ppercent_acc} also shows that using a piecewise linear approximation as described in Section~\ref{sec:tc} for the logistic function does not spoil performance.

\subsection{Fair Training, Certification and Verification}

Figure~\ref{fig:ppercent_fair} shows how the fractions of users with positive outcomes in the two groups ($z=0$ is continuous and $z=1$ is dashed) are gradually balanced as we decrease the fairness constraint~$c$.
These plots can be interpreted as the degree to which disparate impact is mitigated as the constraint is tightened.
The effect is most pronounced for the synthetic dataset by construction.
As discussed above, the collapse for the COMPAS dataset occurs faster than for SLSQP due to the constraints from multiple sensitive attributes.
In the Bank dataset, for large $c$---before the constraint becomes active---the fractions of positive outcomes for $z=1$ differ, which is related to the slightly suboptimal accuracy at large $c$ that needs further investigation.
However, as the constraint becomes active, the fractions are balanced at a similar rate as the baseline.
Overall, our Lagrangian multiplier technique with fixed point numbers and piecewise linear approximations of non-linearities robustly manages to satisfy the p\%-rule proxy at similar rates as the baseline with only minor losses in accuracy on all but the challenging COMPAS dataset.

In Table~\ref{table.timing} we show the online running times of 10 training epochs on a laptop computer.
While training takes several orders of magnitudes longer than a non-MPC implementation, our approach still remains feasible and realistic.
We use the one time offline precomputation of multiplication triples described and timed in~\citet[Table 2]{mohassel2017secureml}.
As pointed out in Section~\ref{sec:mpc}, certification of a trained model requires checking whether $\F(\btheta) > 0$. We already perform this check at least once for each gradient update during training.
It only takes a negligible fraction of the computation time, see Table~\ref{table.timing}.
Similarly, the operations required for certification stay well below one second.

\textbf{Discussion.}
In this section, we have demonstrated the practicability of private and fair model training, certification and verification using MPC as described in Figure~\ref{figure.MPC}.
Using the methods and tricks introduced in Section~\ref{sec:tc}, we can overcome accuracy as well as over- and underflow concerns due to fixed-point numbers.
Offline precomputation combined with a fast C++ implementation yield viable running times for reasonably large datasets on a laptop computer.

\section{Conclusion}
\label{sec:conclusion}

Real world fair learning has suffered from a dilemma:
in order to enforce fairness, sensitive attributes must be examined;
yet in many situations, users may feel uncomfortable in revealing these attributes, or modelers may be legally restricted in collecting and utilizing them.
By introducing recent methods from MPC, and extending them to handle linear constraints as required for various notions of fairness, we have demonstrated that it is practical on real-world datasets to:
(i) certify and sign a model as fair;
(ii) learn a fair model; and
(iii) verify that a fair-certified model has indeed been used;
all while maintaining cryptographic privacy of all users' sensitive attributes.
Connecting concerns in privacy, algorithmic fairness and accountability, our proposal empowers regulators to provide better oversight, modelers to develop fair and private models, and users to retain control over data they consider highly sensitive.

\section*{Acknowledgments}
The authors would like to thank Chris Russell and Phillipp Schoppmann for useful discussions and help with the implementation, as well as the anonymous reviewers for helpful comments.
AG and MK were supported by The Alan Turing Institute under the EPSRC
grant EP/N510129/1.
MV was supported by EPSRC grant EP/M507970/1.
AW acknowledges support from the David MacKay Newton research fellowship at Darwin College, The Alan Turing Institute under EPSRC grant EP/N510129/1 \& TU/B/000074, and the Leverhulme Trust via the CFI.

\bibliographystyle{icml2018}
\bibliography{refs}

\begin{thebibliography}{45}
\providecommand{\natexlab}[1]{#1}
\providecommand{\url}[1]{\texttt{#1}}
\expandafter\ifx\csname urlstyle\endcsname\relax
  \providecommand{\doi}[1]{doi: #1}\else
  \providecommand{\doi}{doi: \begingroup \urlstyle{rm}\Url}\fi

\bibitem[lib()]{liback}
Absentminded crypto kit.
\newblock \url{https://bitbucket.org/jackdoerner/absentminded-crypto-kit}.

\bibitem[Al{-}Rubaie et~al.(2017)Al{-}Rubaie, Wu, Chang, and Kung]{pca}
Al{-}Rubaie, M., Wu, P.~Y., Chang, J.~M., and Kung, S.
\newblock Privacy-preserving {PCA} on horizontally-partitioned data.
\newblock In \emph{{DSC}}, pp.\  280--287. {IEEE}, 2017.

\bibitem[Angwin et~al.(2016)Angwin, Larson, Mattu, and Kirchner]{Angwin2016}
Angwin, J., Larson, J., Mattu, S., and Kirchner, L.
\newblock Machine bias: There is software used across the country to predict
  future criminals. and it is biased against blacks.
\newblock \emph{ProPublica, May}, 23, 2016.

\bibitem[Barocas \& Selbst(2016)Barocas and Selbst]{barocas2016big}
Barocas, S. and Selbst, A.~D.
\newblock Big data's disparate impact.
\newblock \emph{California Law Review}, 104:\penalty0 671--732, 2016.

\bibitem[Bennett~Capers(2012)]{capers2012blind}
Bennett~Capers, I.
\newblock Blind justice.
\newblock \emph{Yale Journal of Law \& Humanities}, 24:\penalty0 179, 2012.

\bibitem[Biddle(2006)]{biddle2006adverse}
Biddle, D.
\newblock \emph{Adverse impact and test validation: A practitioner's guide to
  valid and defensible employment testing}.
\newblock Gower Publishing, Ltd., 2006.

\bibitem[Boyd \& Vandenberghe(2004)Boyd and Vandenberghe]{boydsbook}
Boyd, S. and Vandenberghe, L.
\newblock \emph{Convex optimization}.
\newblock Cambridge university press, 2004.

\bibitem[Calders \& {\v Z}liobait{\.e}(2012)Calders and {\v
  Z}liobait{\.e}]{Calders2012}
Calders, T. and {\v Z}liobait{\.e}, I.
\newblock Why unbiased computational processes can lead to discriminative
  decision procedures.
\newblock In \emph{{Discrimination and Privacy in the Information Society}},
  pp.\  43--59. Springer, 2012.

\bibitem[Damgård et~al.(2012)Damgård, Pastro, Smart, and
  Zakarias]{damgard_multiparty_2012}
Damgård, I., Pastro, V., Smart, N.~P., and Zakarias, S.
\newblock Multiparty computation from somewhat homomorphic encryption.
\newblock In \emph{{CRYPTO}}, volume 7417 of \emph{Lecture Notes in Computer
  Science}, pp.\  643--662. Springer, 2012.

\bibitem[Demmler et~al.(2015{\natexlab{a}})Demmler, Dessouky, Koushanfar,
  Sadeghi, Schneider, and Zeitouni]{DDKSSZ15}
Demmler, D., Dessouky, G., Koushanfar, F., Sadeghi, A., Schneider, T., and
  Zeitouni, S.
\newblock Automated synthesis of optimized circuits for secure computation.
\newblock In \emph{{ACM} Conference on Computer and Communications Security},
  pp.\  1504--1517. {ACM}, 2015{\natexlab{a}}.

\bibitem[Demmler et~al.(2015{\natexlab{b}})Demmler, Schneider, and
  Zohner]{demmler_aby_2015}
Demmler, D., Schneider, T., and Zohner, M.
\newblock {ABY} -- a framework for efficient mixed-protocol secure two-party
  computation.
\newblock In \emph{{NDSS}}. The Internet Society, 2015{\natexlab{b}}.

\bibitem[Dwork et~al.(2012)Dwork, Hardt, Pitassi, Reingold, and Zemel]{Dwork}
Dwork, C., Hardt, M., Pitassi, T., Reingold, O., and Zemel, R.
\newblock Fairness through awareness.
\newblock In \emph{Proceedings of the 3rd Innovations in Theoretical Computer
  Science Conference}, ITCS '12, pp.\  214--226. ACM, 2012.

\bibitem[Faiedh et~al.(2001)Faiedh, Gafsi, and Besbes]{sigmoid-approx}
Faiedh, H., Gafsi, Z., and Besbes, K.
\newblock Digital hardware implementation of sigmoid function and its
  derivative for artificial neural networks.
\newblock \emph{Proceeding of the 13th International Conference on
  Microelectronics, 2001.}, pp.\  189 -- 192, 11 2001.

\bibitem[Feldman et~al.(2015)Feldman, Friedler, Moeller, Scheidegger, and
  Venkatasubramanian]{FeldmanFMSV15}
Feldman, M., Friedler, S., Moeller, J., Scheidegger, C., and
  Venkatasubramanian, S.
\newblock Certifying and removing disparate impact.
\newblock In \emph{Proceedings of the 21th {ACM} {SIGKDD} International
  Conference on Knowledge Discovery and Data Mining}, pp.\  259--268, 2015.

\bibitem[Fredrikson et~al.(2015)Fredrikson, Jha, and
  Ristenpart]{fredrikson2015model}
Fredrikson, M., Jha, S., and Ristenpart, T.
\newblock Model inversion attacks that exploit confidence information and basic
  countermeasures.
\newblock In \emph{Proceedings of the 22nd ACM SIGSAC Conference on Computer
  and Communications Security}, pp.\  1322--1333, 2015.

\bibitem[Friedler et~al.(2016)Friedler, Scheidegger, and
  Venkatasubramanian]{Friedler2016}
Friedler, S.~A., Scheidegger, C., and Venkatasubramanian, S.
\newblock On the (im)possibility of fairness.
\newblock 2016.
\newblock arXiv:1609.07236v1 [cs.CY].

\bibitem[Gascón et~al.(2017)Gascón, Schoppmann, Balle, Raykova, Doerner,
  Zahur, and Evans]{gascon_privacy-preserving_2017}
Gascón, A., Schoppmann, P., Balle, B., Raykova, M., Doerner, J., Zahur, S.,
  and Evans, D.
\newblock Privacy-{Preserving} {Distributed} {Linear} {Regression} on
  {High}-{Dimensional} {Data}.
\newblock \emph{Proceedings on Privacy Enhancing Technologies}, 2017\penalty0
  (4):\penalty0 345--364, October 2017.

\bibitem[Goldreich(2004)]{goldreichbook}
Goldreich, O.
\newblock \emph{The Foundations of Cryptography -- Volume 2, Basic
  Applications}.
\newblock Cambridge University Press, 2004.

\bibitem[Goldreich et~al.(1987)Goldreich, Micali, and
  Wigderson]{DBLP:conf/stoc/GoldreichMW87}
Goldreich, O., Micali, S., and Wigderson, A.
\newblock How to play any mental game or {A} completeness theorem for protocols
  with honest majority.
\newblock In \emph{{STOC}}, pp.\  218--229. {ACM}, 1987.

\bibitem[Graham(2017)]{Graham17}
Graham, C.
\newblock {NHS cyber attack: Everything you need to know about 'biggest
  ransomware' offensive in history}.
\newblock \emph{Telegraph, May 20}, 2017.

\bibitem[Grgi{\'c}-Hla{\v{c}}a et~al.(2018)Grgi{\'c}-Hla{\v{c}}a, Zafar,
  Gummadi, and Weller]{grgic2018}
Grgi{\'c}-Hla{\v{c}}a, N., Zafar, M.~B., Gummadi, K.~P., and Weller, A.
\newblock Beyond distributive fairness in algorithmic decision making: Feature
  selection for procedurally fair learning.
\newblock In \emph{AAAI}, 2018.

\bibitem[Handel et~al.(2014)Handel, Skog, Wahlstrom, Bonawiede, Welch, Ohlsson,
  and Ohlsson]{handel2014insurance}
Handel, P., Skog, I., Wahlstrom, J., Bonawiede, F., Welch, R., Ohlsson, J., and
  Ohlsson, M.
\newblock Insurance telematics: Opportunities and challenges with the
  smartphone solution.
\newblock \emph{IEEE Intelligent Transportation Systems Magazine}, 6\penalty0
  (4):\penalty0 57--70, 2014.

\bibitem[Hardt et~al.(2016)Hardt, Price, and Srebro]{Hardtetal}
Hardt, M., Price, E., and Srebro, N.
\newblock Equality of opportunity in supervised learning.
\newblock In \emph{{NIPS}}, 2016.

\bibitem[Juvekar et~al.(2018)Juvekar, Vaikuntanathan, and
  Chandrakasan]{gazelle}
Juvekar, C., Vaikuntanathan, V., and Chandrakasan, A.
\newblock {G}azelle: A {L}ow {L}atency {F}ramework for {S}ecure {N}eural
  {N}etwork {I}nference.
\newblock \emph{IACR Cryptology ePrint Archive}, 2018:\penalty0 73, 2018.

\bibitem[Keller et~al.(2013)Keller, Scholl, and Smart]{sha-mpc}
Keller, M., Scholl, P., and Smart, N.~P.
\newblock An architecture for practical actively secure {MPC} with dishonest
  majority.
\newblock In \emph{{ACM} Conference on Computer and Communications Security},
  pp.\  549--560. {ACM}, 2013.

\bibitem[Keller et~al.(2017)Keller, Orsini, Rotaru, Scholl, Soria{-}Vazquez,
  and Vivek]{KellerORSSV17}
Keller, M., Orsini, E., Rotaru, D., Scholl, P., Soria{-}Vazquez, E., and Vivek,
  S.
\newblock Faster secure multi-party computation of {AES} and {DES} using lookup
  tables.
\newblock In \emph{{ACNS}}, volume 10355 of \emph{Lecture Notes in Computer
  Science}, pp.\  229--249. Springer, 2017.

\bibitem[Keller et~al.(2018)Keller, Pastro, and
  Rotaru]{DBLP:conf/eurocrypt/KellerPR18}
Keller, M., Pastro, V., and Rotaru, D.
\newblock Overdrive: Making {SPDZ} great again.
\newblock In \emph{{EUROCRYPT} {(3)}}, volume 10822 of \emph{Lecture Notes in
  Computer Science}, pp.\  158--189. Springer, 2018.

\bibitem[Kroll et~al.(2016)Kroll, Huey, Barocas, Felten, Reidenberg, Robinson,
  and Yu]{kroll2016accountable}
Kroll, J.~A., Huey, J., Barocas, S., Felten, E.~W., Reidenberg, J.~R.,
  Robinson, D.~G., and Yu, H.
\newblock Accountable algorithms.
\newblock \emph{University of Pennsylvania Law Review}, 165, 2016.

\bibitem[Lichman(2013)]{uci}
Lichman, M.
\newblock {UCI} machine learning repository, 2013.
\newblock URL \url{http://archive.ics.uci.edu/ml}.

\bibitem[Lindell(2016)]{lindell_how_2016}
Lindell, Y.
\newblock How {To} {Simulate} {It} -- {A} {Tutorial} on the {Simulation}
  {Proof} {Technique}.
\newblock \emph{IACR Cryptology ePrint Archive}, 2016:\penalty0 46, 2016.

\bibitem[Liu et~al.(2017)Liu, Juuti, Lu, and Asokan]{minionn}
Liu, J., Juuti, M., Lu, Y., and Asokan, N.
\newblock Oblivious neural network predictions via minionn transformations.
\newblock In \emph{{CCS}}, pp.\  619--631. {ACM}, 2017.

\bibitem[Mohassel \& Zhang(2017)Mohassel and Zhang]{mohassel2017secureml}
Mohassel, P. and Zhang, Y.
\newblock {SecureML: A system for scalable privacy-preserving machine
  learning}.
\newblock In \emph{IEEE Symposium on Security and Privacy (SP)}, pp.\  19--38,
  2017.

\bibitem[Nikolaenko et~al.(2013{\natexlab{a}})Nikolaenko, Ioannidis, Weinsberg,
  Joye, Taft, and Boneh]{nikolaenko_matrix-factorisation}
Nikolaenko, V., Ioannidis, S., Weinsberg, U., Joye, M., Taft, N., and Boneh, D.
\newblock Privacy-preserving matrix factorization.
\newblock In \emph{{ACM} Conference on Computer and Communications Security},
  pp.\  801--812. {ACM}, 2013{\natexlab{a}}.

\bibitem[Nikolaenko et~al.(2013{\natexlab{b}})Nikolaenko, Weinsberg, Ioannidis,
  Joye, Boneh, and Taft]{nikolaenko_privacy-preserving_2013}
Nikolaenko, V., Weinsberg, U., Ioannidis, S., Joye, M., Boneh, D., and Taft, N.
\newblock Privacy-preserving ridge regression on hundreds of millions of
  records.
\newblock In \emph{{IEEE} Symposium on Security and Privacy}, pp.\  334--348.
  {IEEE} Computer Society, 2013{\natexlab{b}}.

\bibitem[Schreurs et~al.(2008)Schreurs, Hildebrandt, Kindt, and
  Vanfleteren]{Schreurs2008}
Schreurs, W., Hildebrandt, M., Kindt, E., and Vanfleteren, M.
\newblock {\textit{Cogitas, Ergo Sum}. The Role of Data Protection Law and
  Non-discrimination Law in Group Profiling in the Private Sector}.
\newblock In \emph{Profiling the European Citizen}, pp.\  241--270. Springer,
  2008.

\bibitem[Tram{\`e}r et~al.(2016)Tram{\`e}r, Zhang, Juels, Reiter, and
  Ristenpart]{tramer2016stealing}
Tram{\`e}r, F., Zhang, F., Juels, A., Reiter, M.~K., and Ristenpart, T.
\newblock Stealing machine learning models via prediction apis.
\newblock In \emph{USENIX Security Symposium}, pp.\  601--618, 2016.

\bibitem[Veale \& Binns(2017)Veale and Binns]{VealeBinns2017}
Veale, M. and Binns, R.
\newblock Fairer machine learning in the real world: Mitigating discrimination
  without collecting sensitive data.
\newblock \emph{Big Data \& Society}, 4\penalty0 (2), 2017.

\bibitem[Veale \& Edwards(2018)Veale and Edwards]{vealeedwardsa29}
Veale, M. and Edwards, L.
\newblock {Clarity, Surprises, and Further Questions in the Article 29 Working
  Party Draft Guidance on Automated Decision-Making and Profiling}.
\newblock \emph{Computer Law \& Security Review}, 2018.
\newblock \doi{10.1016/j.clsr.2017.12.002}.

\bibitem[Yao(1986)]{yao_how_1986}
Yao, A. C.-C.
\newblock How to {Generate} and {Exchange} {Secrets} ({Extended} {Abstract}).
\newblock In \emph{{FOCS}}, pp.\  162--167. IEEE Computer Society, 1986.

\bibitem[Zafar et~al.(2017{\natexlab{a}})Zafar, Valera, Gomez{-}Rodriguez, and
  Gummadi]{Zafaretal}
Zafar, M.~B., Valera, I., Gomez{-}Rodriguez, M., and Gummadi, K.~P.
\newblock Fairness beyond disparate treatment {\&} disparate impact: Learning
  classification without disparate mistreatment.
\newblock In \emph{WWW}, 2017{\natexlab{a}}.

\bibitem[Zafar et~al.(2017{\natexlab{b}})Zafar, Valera, Rodriguez, Gummadi, and
  Weller]{zafar2017parity}
Zafar, M.~B., Valera, I., Rodriguez, M., Gummadi, K., and Weller, A.
\newblock From parity to preference-based notions of fairness in
  classification.
\newblock In \emph{Advances in Neural Information Processing Systems}, pp.\
  228--238, 2017{\natexlab{b}}.

\bibitem[Zafar et~al.(2017{\natexlab{c}})Zafar, Valera, Rodriguez, and
  Gummadi]{zafar_fairness}
Zafar, M.~B., Valera, I., Rodriguez, M.~G., and Gummadi, K.~P.
\newblock {Fairness Constraints: Mechanisms for Fair Classification}.
\newblock In \emph{{AISTATS}}, 2017{\natexlab{c}}.

\bibitem[Zahur \& Evans(2015{\natexlab{a}})Zahur and Evans]{zahur2015obliv}
Zahur, S. and Evans, D.
\newblock Obliv-c: A language for extensible data-oblivious computation.
\newblock \emph{IACR Cryptology ePrint Archive}, 2015:\penalty0 1153,
  2015{\natexlab{a}}.

\bibitem[Zahur \& Evans(2015{\natexlab{b}})Zahur and
  Evans]{zahur_obliv-c:_2015}
Zahur, S. and Evans, D.
\newblock Obliv-{C}: {A} {Language} for {Extensible} {Data}-{Oblivious}
  {Computation}.
\newblock \emph{IACR Cryptology ePrint Archive}, 2015:\penalty0 1153,
  2015{\natexlab{b}}.

\bibitem[{\v{Z}}liobait{\.{e}} \& Custers(2016){\v{Z}}liobait{\.{e}} and
  Custers]{Zliobaite2016}
{\v{Z}}liobait{\.{e}}, I. and Custers, B.
\newblock Using sensitive personal data may be necessary for avoiding
  discrimination in data-driven decision models.
\newblock \emph{Artificial Intelligence and Law}, 24\penalty0 (2):\penalty0
  183--201, 2016.

\end{thebibliography}

\clearpage
\appendix

\section{Details of the MPC Protocols}
\label{sec:mpcdetails}

\paragraph{Secret sharing.}
A secret sharing scheme allows one to split a value $x$ (the secret) among two parties, so that no party has unilateral access to $x$.
In our setting, a user Alice will secret share a sensitive value, for example her race, among a modeler~$\dc$ and a regulator~$\reg$. Several secret sharing schemes exist, including \emph{Shamir secret sharing}, \emph{xor sharing}, \emph{Yao sharing}, or \emph{arithmetic multiplicative/additive sharing}.
In this work we alternate between Yao sharing and additive sharing for efficiency. In the latter, the value~$x$ is represented in a finite domain~$\bZ_q$ with, for example~$q=2^{32}$.
To share her race, Alice samples a value $r$ from~$\bZ_q$ uniformly at random, and sends $x - r$ to~$\dc$ and~$r$ to $\reg$.
We call each of $x-r$ and~$r$ a \emph{share}, and denote them as~$\share{x}{1}$ and~$\share{x}{2}$.
Now~$\dc$ and~$\reg$ can recover~$x$ by adding their shares, but each share on its own does not reveal anything about the value of~$x$ (other than that it is smaller than~$q$). Note that the case where $q=2$ corresponds to xor sharing.

\paragraph{Function evaluation.}
MPC can be classified in two groups depending on how~$f$ is represented:
either as a Boolean or arithmetic circuit. All
protocols proceed by having the parties jointly evaluate the circuit,
processing it gate by gate.
For each gate~$g$ for which the value for the input wires~$x, y$
is shared among the parties, the parties run a subprotocol to produce the value
$z = g(x,y)$
of the output wire, again shared,
without revealing any information in the process.
In the setting where we use arithmetic additive sharing,
the two parties $\dc$ and $\reg$ hold shares,
$\share{x}{1}$,$\share{y}{1}$ and $\share{x}{2}$,$\share{y}{2}$,
respectively. In this case, $f$ is represented as
an arithmetic circuit, and hence each gate $g$ in the circuit
is either an addition or a multiplication.
Note that if $g$ is an addition gate, then
a sharing of $z = g(x, y)$ can be obtained
by having each party simply compute locally, i.e.,
without any interaction, $\share{z}{i} = \share{x}{i} + \share{y}{i}$,
for $i\in\{1, 2\}$. If $g$ is a multiplication, the
subprotocol to compute shares of $z$ is much more costly.
Fortunately, it can be divided into an offline and an online phase.

\paragraph{The preprocessing model in MPC.}
In this model, two parties $P_1, P_2$ engage in an offline phase, which is data independent,
and compute (and store) \emph{shared multiplication triples} of the form $(a, b, c)$, with $c = ab$.
Here, $a,b\in\F_q$ are drawn uniformly at random, and each value $a, b, c$ is shared
among the parties as explained above. In the online phase, a multiplication gate
$z = \mul(x, y)$ on shared values $x, y$ can be evaluated as follows:
(1) each $P_i$ sets $\share{e}{i} = \share{x}{i} - \share{a}{i}$ and
$\share{f}{i} = \share{y}{i} - \share{b}{i}$,
(2) the parties exchange their shares of $e$ and $f$ and reconstruct these values locally, and
(3) each $P_i$ computes $\share{z}{i} = (i-1)ef + f\share{a}{i} + e\share{b}{i} + \share{c}{i}$.
The correctness of this protocol can be easily checked. Privacy relies on the uniform randomness of
$a, b$, and hence $\share{e}{i}$ and $\share{f}{i}$ completely
mask the values of $\share{x}{i}$ and $\share{y}{i}$, respectively.
For a formal proof see~\cite{demmler_aby_2015}.

Hence, for each multiplication in the function to be evaluated, the
parties need to jointly generate a multiplication triple in advance.
For computations with many multiplications (like in our case) this can be a costly
process. However, this constraint is easy to accommodate in our architecture
for private fair model training, as $\dc$ and $\reg$ can run the offline phase
once ``overnight''.
Arithmetic multiplication via precomputed triples is a common technique, used in several
popular MPC frameworks~\cite{demmler_aby_2015, damgard_multiparty_2012}.
In this setting, several protocols for triple generation (which we did not describe)
are available~\cite{DBLP:conf/eurocrypt/KellerPR18}, and under continuous improvement.
These protocols are often based on either Oblivious Transfer or Homomorphic Encryption.

\paragraph{The two-server model for multi-party learning.}
Due to a sequence of theoretical and engineering breakthroughs,
in the last three decades MPC has gone from being a mathematical curiosity
to a technology of practical interest with commercial applications.
Several generic protocols for MPC exists, such as the
ones based on arithmetic sharing~\cite{damgard_multiparty_2012},
garbled circuits~\cite{yao_how_1986}, or GMW~\cite{DBLP:conf/stoc/GoldreichMW87},
with several available implementations~\cite{demmler_aby_2015, zahur_obliv-c:_2015}.
These protocols have different trade-offs in terms
of the number of parties they support, network requirements,
and scalability for different kinds of computations.
In our work, we focus on the $2$-party case, as the MPC computation is done by
$\dc$ and $\reg$. The idea of privately outsourcing computation to two non-colluding parties in this way is recurrent in MPC, and often referred to as the two-server model~\cite{mohassel2017secureml, gascon_privacy-preserving_2017, nikolaenko_privacy-preserving_2013, pca}.

While generic protocols exist, these do not yet scale to input sizes typically encountered in machine learning applications like ours. To circumvent this limitation, techniques tailored to specific applications have been proposed. Our protocols fall in this category, extending the SGD protocol from~\cite{mohassel2017secureml}, in which the following useful accelerating techniques are presented.
\begin{itemize}
\item Efficient rescaling: As our arithmetic shares represent fixed-point numbers, we need to rescale by the precision $p$ after every multiplication. This involves dividing by $2^p$, an expensive operation to do in MPC, and in particular in arithmetic sharing. Mohassel et al.\ show an elegant solution to this problem: the parties can rescale locally by dropping $p$ bits of their shares. It is not hard to see that this might produce the wrong result. However, the parameters of the arithmetic secret sharing scheme can be set such that with a tunable arbitrarily large probability the error is at most $\pm 1$. This trick can be used for any division by a power of two.
\item Alternating sharing types: As already pointed out in previous work~\cite{demmler_aby_2015}, alternating between secret sharing schemes can provide significant acceleration for some applications. Intuitively, arithmetic operations are fast in arithmetic shares, while comparisons are fast in schemes that represent functions as Boolean circuits. Examples of the latter are the GMW protocol and Yao's garbled circuits. In our implementation, we follow this recipe and implement matrix-vector multiplication using arithmetic sharing, while for evaluating our variant of sigmoid, we rely on the protocol from~\cite{mohassel2017secureml} implemented with garbled circuits using the Obliv-C framework~\cite{zahur_obliv-c:_2015}.
\item Matrix multiplication triples: Another observation made by Mohassel et al.\ is that the idea described above for preprocessing multiplications over arithmetic shares can be reinterpreted at the level of matrices. This results in a faster online and offline phase (see~\cite{mohassel2017secureml} for details).
\end{itemize}

\paragraph{How to prove that a protocol is secure.}
We did not provide a formal definition of security in this paper, and instead referred the reader to~\cite{mohassel2017secureml}. In MPC, privacy in the case of semi-honest adversaries is argued in the simulation paradigm (see~\cite{goldreichbook} or~\cite{lindell_how_2016} for formal definitions and detailed proofs). Intuitively, in this paradigm one proves that every inference that a party---in our case either $\reg$ or $\dc$---could draw from observing the execution trace of the protocol could also be drawn from the output of the execution and the party's input. This is done by proving the existence of a {\em simulator} that can produce an execution trace that is indistinguishable from the actual execution trace of the protocol. A crucial point is that the simulator only has access to the input and output of the party being simulated.

\section{Details of Fair Model Training}
\label{sec:algo}

\subsection{The Fair Training Algorithm}

Algorithm~\ref{algo:lagrange} describes the computations $\dc$ and $\reg$ have to perform for fair model training using the Lagrangian multiplier technique and the $p\%$-rule from eq.~\eqref{eq:ppercent}.
In the next subsection we describe the parameter values.
We implicitly assume all computations are performed jointly on additively shared secrets by $\dc$ and $\reg$ as described in Section~\ref{sec:mpc}.
This means that $\dc$ and $\reg$ each receive a secret share of the protected attributes~$\Z$.
Following the protocols outlined in Section~\ref{sec:mpc}, they can then jointly evaluate the steps in Algorithm~\ref{algo:lagrange}.
This allows them to operate on the sensitive values within the MPC computation, while preventing unilateral access to them by $\dc$ and $\reg$.
The result of these computations is the same as evaluating the algorithm as described with data in the clear.

\textsc{BlockedMultShiftAvg} stands for the blocked matrix multiplication to avoid overflow for fixed-point numbers described towards the end of Section~\ref{sec:tc}.
Note that it already contains the division by~$n$.
The averaging within the blocked matrix multiplications as well as over the results thereof are done by fast bit shifts instead of slow MPC division circuits.
This is possible, because we chose all parameters such that divisions are always by powers of two.

\begin{figure}
\centering
\includegraphics{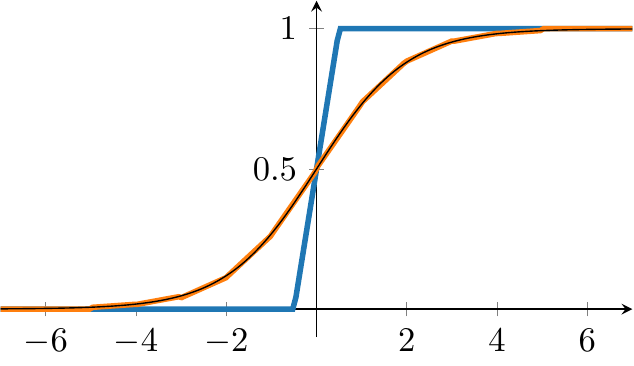}
\caption{Piecewise linear approximations for the non-linear sigmoid function (in black) from \citet{mohassel2017secureml} in blue and from \citet{sigmoid-approx} in orange.}
\label{fig:approximations}
\end{figure}

We found the piecewise linear approximation of the sigmoid function introduced in~\cite{mohassel2017secureml}
\begin{equation*}
\textsc{SigmoidApprox}(x) :=
\begin{cases}
0 &\text{if } x \le -\frac{1}{2}\:, \\
x + \frac{1}{2} &\text{if } -\frac{1}{2} < x < \frac{1}{2}\:, \\
1 &\text{if } x \ge \frac{1}{2}\:.
\end{cases}
\end{equation*}
to work best, see Figure~\ref{fig:approximations}.

\begin{algorithm}[htb]
  \caption{Fair model training with private sensitive values using Lagrangian multipliers for~$\F(\btheta) = \nicefrac{1}{n} |\Z^{\top} X| - \const$.}
  {\bf Parties:}~~$\dc$, $\reg$.\\
  {\bf Input:}~~($\dc$) $\mat{\share{Z}{1}} \in \mathbb{Z}_q^{n\times p}$\\
  {\bf Input:}~~($\reg$) $\mat{X} \in \mathbb{Z}_q^{n\times d}$, $\vec{y} \in \mathbb{Z}_q^n$, $\mat{\share{Z}{2}} \in \mathbb{Z}_q^{n\times p}$\\
  {\bf Input:}~~(Public) Learning rates $\eta_{\btheta}, \eta_{\blambda}$, number of training examples~$n$, minibatch size $2^s$, constraints $\vec{c}\in \mathbb{Z}_q^p$, number of epochs $N_e$.

  \begin{algorithmic}[1]
    \STATE $\btheta \gets \vec{0}$, $\blambda \gets \vec{0}$

    \STATE $\A \gets \textsc{BlockedMultShiftAvg}(\Z^{\top}, \X)$

    \FORALL{$j$ from $1$ to $N_e$}

    \FORALL{$i$ from $1$ to $\nicefrac{n}{2^s}$}

    \STATE $(\X_{i},\y_{i}) \gets \textsc{SampleMinibatch}(\X, \y)$

    \STATE $\F \gets |\A \btheta| - c$

    \STATE $\nabla_{\blambda} \gets \max\{\F, \vec{0} \}$

    \STATE $\sigma \gets \textsc{SigmoidApprox}(\X_i \btheta)$

    \STATE $\nabla_{\btheta}^{\mathrm{BCE}} \gets \textsc{ShiftDivide}(\X_i^{\top}(\sigma - \y_i), 2^s)$

    \STATE $\nabla_{\btheta}^{\mathrm{CON}} \gets
      \begin{cases}
      \A^{\top} \blambda, &\text{if } \A > \vec{0} \wedge \F > 0 \\
      - \A^{\top} \blambda, &\text{if } \A < \vec{0} \wedge \F > 0\\
      \vec{0}, &\text{if } \F \le 0
      \end{cases}$

    \STATE $\btheta \gets \btheta - \eta_{\btheta} (\xi^{\mathrm{BCE}}_j \nabla_{\btheta}^{\mathrm{BCE}} + \xi^{\mathrm{CON}}_j \nabla_{\btheta}^{\mathrm{CON}})$

    \STATE $\blambda \gets \max\{\blambda + \eta_{\blambda} \nabla_{\blambda}, \vec{0}\}$
    \ENDFOR
    \ENDFOR
  \end{algorithmic}

  {\bf Output:}~~Parameters~$\btheta$
  \label{algo:lagrange}
\end{algorithm}

\subsection{Description of Training Parameters}

All our experiments use a batch size of~64, a fixed number of epochs scaling inversely with dataset size~$n$ (such that we always perform roughly 15\,000 gradient updates), fixed learning rates of~$\eta_{\btheta} = 10^{-4}, \eta_{\blambda} = 0.05$, and an annealing schedule for~$\nicefrac{1}{t}$ in the interior point logarithmic barrier method as described by \citet{boydsbook}.
The weights for the gradients of the regular binary cross entropy loss (BCE) and the loss from the constraint terms (CON) follow the schedules
\begin{equation*}
\xi^{\mathrm{BCE}}_j = \frac{N_e}{N_e + j}\:,
\qquad
\xi^{\mathrm{CON}}_j = \frac{N_e + 10 j}{N_e}\:.
\end{equation*}
Weight decay, adaptive learning rate schedules. and momentum neither consistently improved nor impaired training.
Therefore, all reported numbers were achieved with vanilla SGD, for fixed learning rates, and without any regularization.
After extensive testing on all datasets, we converged to a fixed-point representation with 16 bits for the integer and fractional part respectively.
The smaller the number of bits, the faster the MPC implementation and the higher the risk of loss of precision or over- and underflows.
We found~16 bits to be the minimally needed precision for all our experiments to work.

\section{Additional Experimental Results}
\label{sec:additional}

\subsection{Results on Remaining Datasets}

\begin{figure*}
\centering
\includegraphics{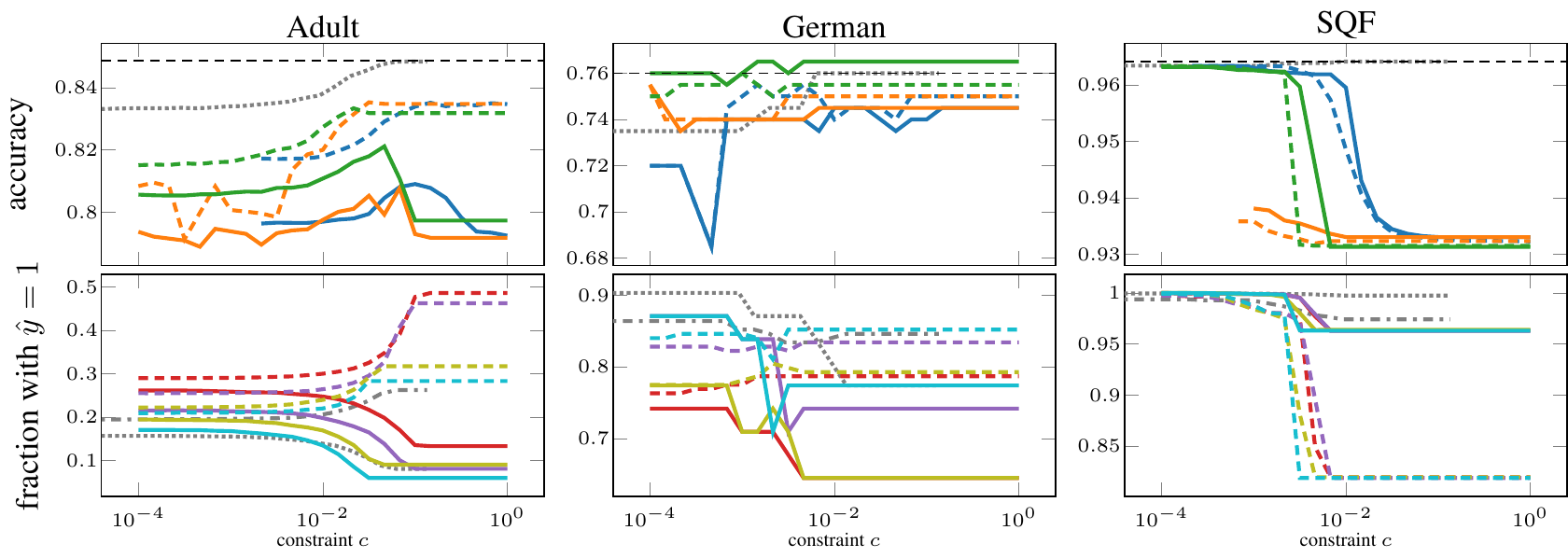}
\vspace{-0.7cm}
\caption{\textbf{First row:} The color code is {\color{py-3-1} blue: iplb}, {\color{py-3-2} orange: projected}, {\color{py-3-3} green: Lagrange} with \emph{continuous} lines for no approximation and \emph{dashed} lines for piecewise linear approximation.
The gray dotted line is the baseline and the dashed black line marks unconstrained logistic regression.
\textbf{Second row:} \emph{Continuous/dotted} lines correspond to $z=0$ and \emph{dashed/dash-dotted} lines to $z=1$. The color code is
({\color{py-4-1}red: no approx. + float},
{\color{py-4-2}purple: no approx. + fixed},
{\color{py-4-3}yellow: pw linear + float},
{\color{py-4-4}turquoise: pw linear + fixed},
gray: baseline).}
\label{fig:ppercent_rest_accs}
\vspace{-0.2cm}
\end{figure*}

Analogously to Figures~\ref{fig:ppercent_acc} and Figure~\ref{fig:ppercent_fair} we report the results on test accuracy as well as the mitigation of disparate impact for the Lagrangian multiplier method in Figure~\ref{fig:ppercent_rest_accs}.
In the Adult dataset we are able to mitigate disparate impact with slightly worse accuracy as compared to the baseline.
Note that the German dataset contains only 512 training and 200 test examples, which explains the discrete jumps in accuracy in minimal steps of $\nicefrac{1}{200} = 0.005$.
Hence, even though the Lagrangian multiplier technique here consistently removes disparate impact to a similar extent as the baseline, interpretations of results on such small datasets require great care.
For the much larger stop, question and frisk dataset we again observe the curious initial increase in accuracy similar to our observations for the Bank dataset.
In this dataset about $93\%$ of all samples have positive labels, which explains the near optimal accuracy when collapsing to always predict 1, which happens for the baseline as well for our method at a similar rate as $c$ decreases.

\begin{figure*}
\centering
\includegraphics{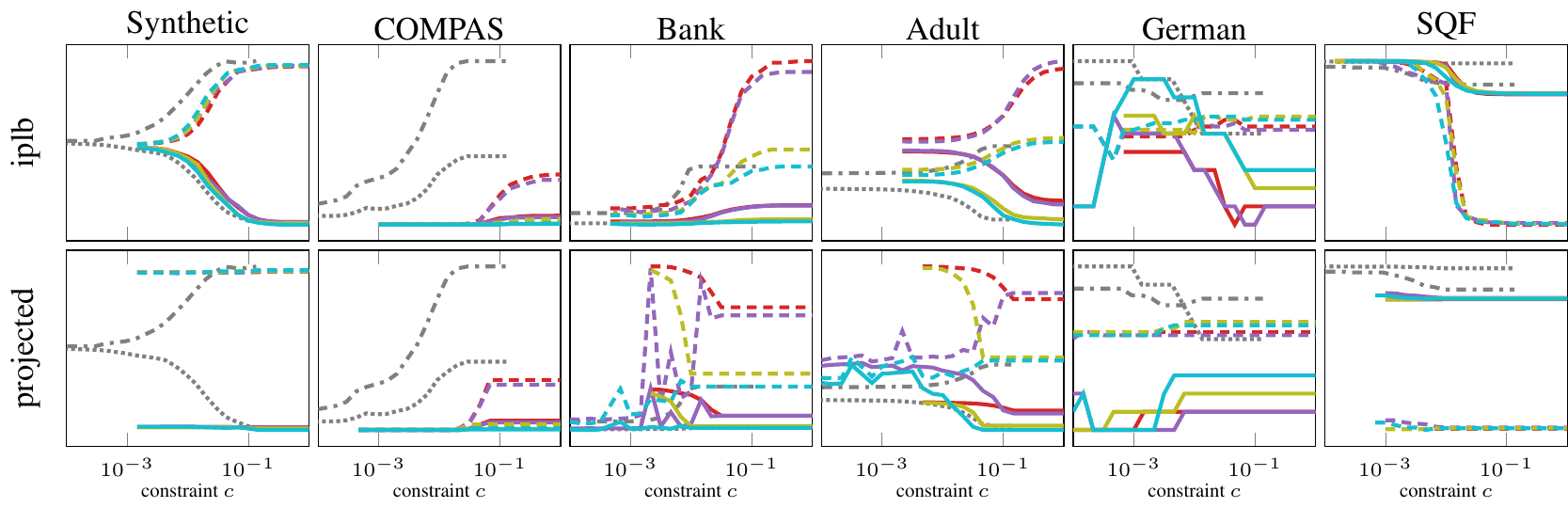}
\vspace{-0.2cm}
\caption{We plot the fraction of people with $z=0$ (\emph{continuous/dotted}) and with $z=1$ (\emph{dashed/dash-dotted}) who get assigned positive outcomes over the constraint $c$ for 5 different datasets. The different colors correspond to
({\color{py-4-1}red: no approximation + floats},
{\color{py-4-2}purple: no approximation + fixed-point},
{\color{py-4-3}yellow: piecewise linear + floats},
{\color{py-4-4}turquoise: piecewise linear + fixed-point},
gray: baseline).}
\label{fig:iplb_and_projected}
\vspace{-0.6cm}
\end{figure*}

\subsection{Disadvantages of Other Optimization Methods}

In Section~\ref{sec:experiments} we suggest the Lagrangian multiplier technique for fair model training using fixed-point numbers.
Here we substantiate this suggestion with further empirical evidence.
Figure~\ref{fig:iplb_and_projected} shows analogous results to Figure~\ref{fig:ppercent_fair} and the second row of Figure~\ref{fig:ppercent_rest_accs}.
These plots reveal the shortcomings of the interior point logarithmic barrier and the projected gradient methods.

\paragraph{Interior Point Logarithmic Barrier method.}
While the interior point logarithmic barrier method does balance the fractions of people being assigned positive outcomes between the two different demographic groups when the constraint is tightened, it soon breaks down entirely due to overflow and underflow errors.
The number of failed runs was substantially higher than for the Lagrangian multiplier technique.
As explained in~\citep{boydsbook}, when we increase the parameter~$t$ of the interior point logarithmic barrier method during training, the barrier becomes steeper, approaching the function
\begin{equation*}
I_{-}(x) =
\begin{cases}
    0 & \text{for } x \le 0\:, \\
    \infty & \text{for } x > 0\:.
  \end{cases}
\end{equation*}
From this it becomes obvious that when facing tight constraints, the gradients might change from almost zero to extremely large values within a single update of the parameters~$\btheta$.
Moreover, iplb requires careful tuning and scheduling of~$t$.
Hence, the interior point logarithmic barrier method, while achieving good results over some domains, is not well suited for MPC.

\paragraph{Projected gradient method.}
In Figure~\ref{fig:iplb_and_projected}, we observe that the projected gradient method seems to fail in most cases, since it does not actually balance the fractions of positive outcomes across the sensitive groups.
There is a simple explanation why it can satisfy the constraint~$\F(\btheta) \le 0$ for the $p\%$-rule even with small~$\const$ and still retain near optimal accuracy.
Note that the accuracy only depends on the direction of~$\btheta$, i.e., it is invariant to arbitrary rescaling of~$\btheta$.
Since the constraint~$\F(\btheta) = |\A \btheta | - \const \le 0$ is always satisfied for $\btheta = 0$, dividing any $\btheta$ by a large enough factor will result in a classifier that achieves equal accuracy and satisfies the constraint (by continuity).
However, minimizing the loss in the original logistic regression optimization problem (or equivalently maximizing the likelihood), which is not invariant under rescaling of $\btheta$, counteracts shrinking $\btheta$ as it enforces high confidence of decisions, i.e., large~$\btheta$.
The projection method produces high accuracy classifiers with small weights that formally fulfill the fairness constraint, but do not properly mitigate disparate impact as measured by the true $p$\%-rule instead of the computational proxy.
It also often fails for small constraint values, as the projection matrix in eq.~\eqref{eq:projected} turns out to become near singular producing over- and underflow errors.

\section{Clarification of Privacy or Secrecy}

In this work, privacy or secrecy constraints are separate from other theorized, setup-dependent attacks, e.g., model extraction~\cite{tramer2016stealing} or inversion~\cite{fredrikson2015model}. If relevant, modelers may need to consider these separately.

\end{document}